\documentclass[10pt,twocolumn,letterpaper]{article} 

\usepackage{avss}
\usepackage{times}
\usepackage{epsfig}
\usepackage{graphicx}
\usepackage{amsmath}
\usepackage{amssymb}
\usepackage{enumitem}
\usepackage{subfigure}
\usepackage{caption}
\usepackage{threeparttable} 
\usepackage{authblk} % for multiple authors with multiple affiliations
\usepackage{amssymb}% http://ctan.org/pkg/amssymb
\usepackage{pifont}% http://ctan.org/pkg/pifont
\newcommand{\cmark}{\ding{51}}%
\newcommand{\xmark}{\ding{55}}%
\usepackage{multirow}
\usepackage{url}

\avssfinalcopy % *** Uncomment this line for the final submission

\def\avssPaperID{17} % *** Enter the AVSS Paper ID here

\ifavssfinal\fi
\begin{document}
%%%%%%%%% TITLE
% \title{Accident Forecasting in CCTV Traffic camera videos}
\title{CADP:~A Novel Dataset for CCTV Traffic Camera based Accident Analysis}
% \author{First Author\\
% Institution1\\
% Institution1 address\\
% {\tt\small firstauthor@i1.org}
% % For a paper whose authors are all at the same institution, 
% % omit the following lines up until the closing ``}''.
% % Additional authors and addresses can be added with ``\and'', 
% % just like the second author.
% % To save space, use either the email address or home page, not both
% \and
% Second Author\\
% Institution2\\
% First line of institution2 address\\
% {\tt\small secondauthor@i1.org}
% }
\author[1,*]{Ankit Parag Shah}
\author[1,*]{Jean-Bapstite Lamare}
\author[2,*]{Tuan Nguyen-Anh}
\author[1]{Alexander Hauptmann}
\affil[1]{Language Technology Institute, Carnegie Mellon University, USA}
\affil[2]{University of Tokyo, Japan}
\affil[*]{denotes shared first authorship}
\maketitle
% \thispagestyle{empty}

%%%%%%%%% ABSTRACT
\begin{abstract}
		This paper presents a novel dataset for traffic accidents analysis.
		Our goal is to resolve the lack of public data for research about automatic spatio-temporal annotations for traffic safety in the roads.
		Through the analysis of the proposed dataset, we observed a significant degradation of object detection in pedestrian category in our dataset, due to the object sizes and complexity of the scenes.
		To this end, we propose to integrate contextual information into conventional Faster R-CNN using \textit{Context Mining~(CM)} and \textit{Augmented Context Mining~(ACM)} to complement the accuracy for small pedestrian detection.
		Our experiments indicate a considerable improvement in object detection accuracy:~+8.51\% for CM and~+6.20\% for ACM.
		Finally, we demonstrate the performance of accident forecasting in our dataset using Faster R-CNN and an Accident LSTM architecture.
		We achieved an average of 1.684 seconds in terms of Time-To-Accident measure with an Average Precision of 47.25\%. Our  Webpage  for  the  paper  is \url{https://goo.gl/cqK2wE}
\end{abstract}

%%%%%%%%% BODY TEXT
\section{Introduction}
%\blfootnote{978-1-5386-9294-3/18/\$31.00~\textcopyright2018 IEEE}
% Please follow the steps outlined below when submitting your manuscript to
% the IEEE Computer Society Press.  This style guide now has several
% important modifications (for example, you are no longer warned against the
% use of sticky tape to attach your artwork to the paper), so all authors
% should read this new version.
	\begin{figure}[tb]
		\centering
		\subfigure[]{\includegraphics[clip,width=\columnwidth]{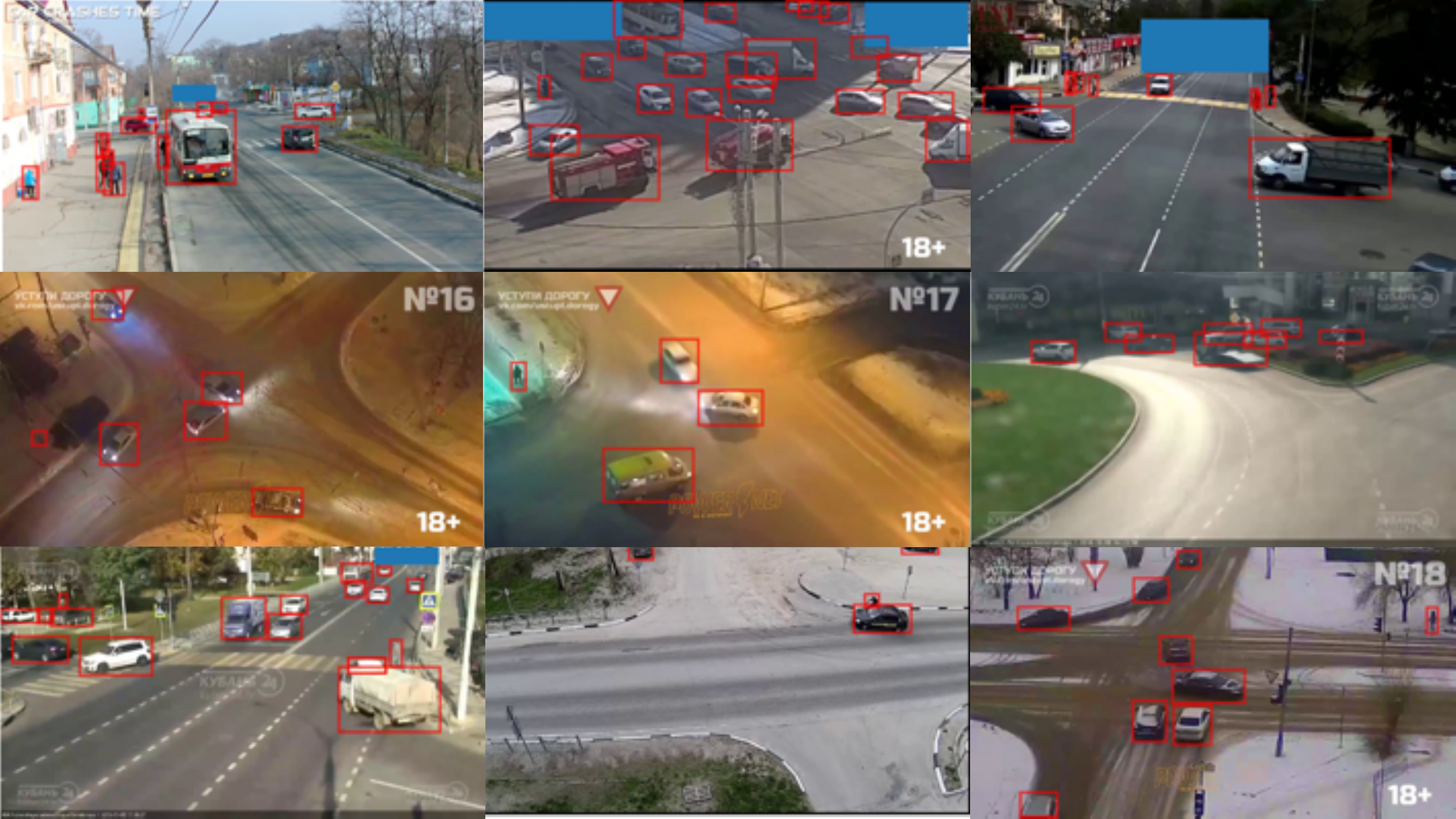}}
	    \subfigure[]{\includegraphics[clip,width=\columnwidth]{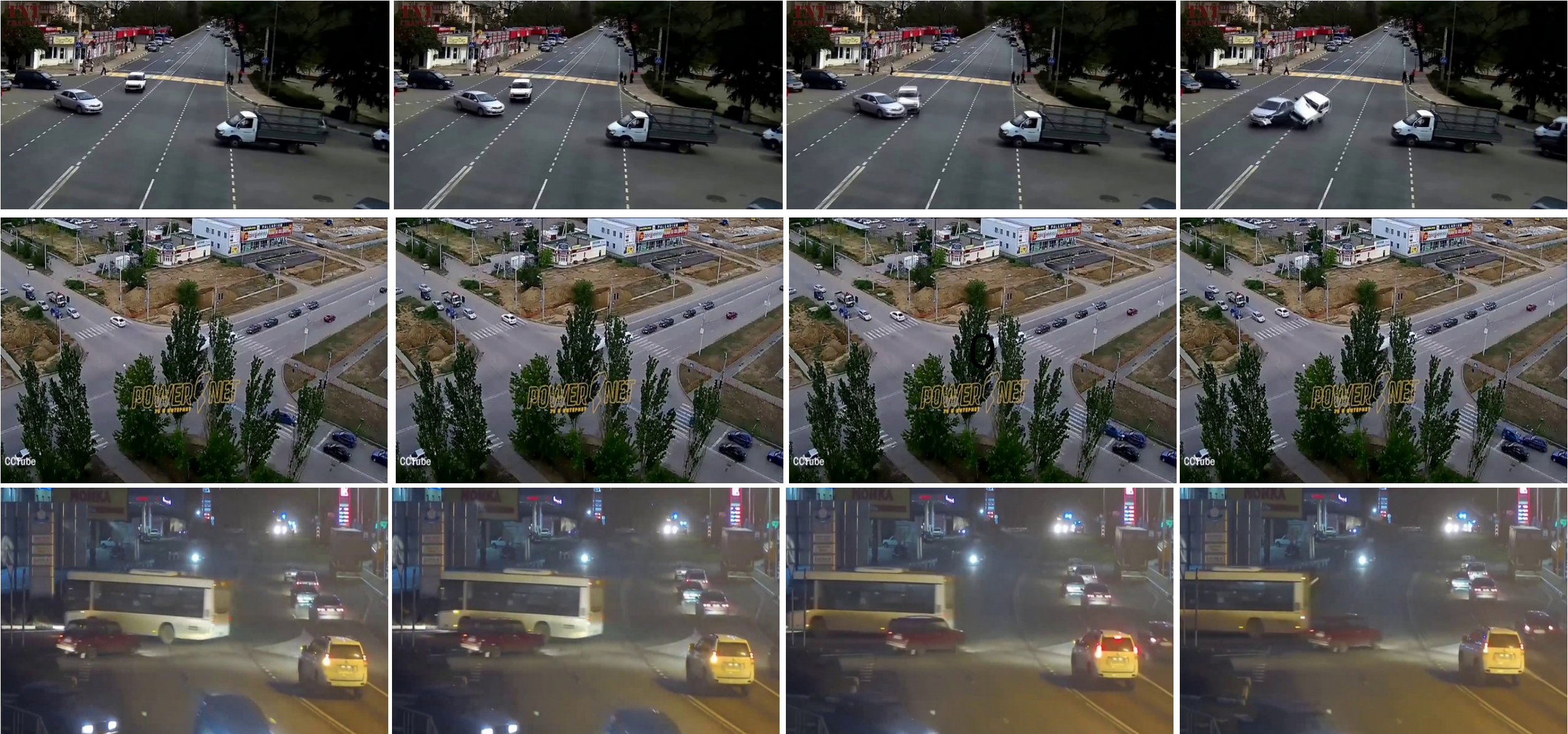}}
		\caption{\textit{(a)~Can you depict \textbf{where} the accidents happen in the image plane?;~(b)~Can you \textbf{identify} and \textbf{forecast} the sequences containing accidents? }~Best viewed in color.}
		\label{fig1}
	\end{figure}
	% Problem and it importance
	According to the National Safety Council, an estimated 40,200 people died on the nation's roads in 2016, making motor vehicle crashes the second leading cause of unintentional deaths in the United States~\cite{nationalreport1}.
	Our work is devoted toward the objective of making roads safer with neutral views of the accidents from traffic cameras which are installed high on a corner of the road. 
	The advantage of third-person views over first-person views is two-fold:~(i)~third-person views have a fixed and wider view because they are mounted higher; and~(ii)~traffic camera views can be used in the public for a vast amount of vehicles daily, thus, the cost per vehicle per day is lower.
	While the former enhances the views of traffic accidents, the latter enhances the trade-offs between cost and safety:~higher quality (HD 720p-1080p) and better featured cameras, such as Palt-Tilt-Zoom HD cameras, can be used with low cost to monitor the public crowds.
	Although the exploitation of traffic camera views is promising, the number of datasets aimed at learning to detect and predict the accidents on those views is limited due to several unaffordable factors:~(i)~traffic accidents are rare events, thus, acquiring enough data by recording at a road intersection is infeasible because one may have to wait endlessly for the an accident to happen; and~(ii)~the access to traffic camera data is legally difficult to obtain in practice.
	To this end, we propose an effective data collection process to exploit the edge-case data:~YouTube videos of traffic accidents that have been uploaded by users over the world.
	We exploited the search engine of YouTube, and added our annotation processes using both internal annotators and outside workers to build a novel dataset, the Car Accident Detection and Prediction~(CADP) dataset for multiple purposes:~temporal segmentation, object detection, tracking, vehicle collision, accident detection and prediction.
	Our dataset contains 230 videos, each video containing at least one accident captured from fixed traffic camera views and 1,416 segments of traffic accidents.
	Moreover, we selected 205 segments with HD quality to annotate spatio-temporal data for object detection, tracking and collision detection. The data is made available for research use available through \url{https://goo.gl/cqK2wE}. 

	% Solutions and contributions
	\noindent\textbf{Contributions}~Our contributions are as follows:\vspace{-2mm}
	\begin{itemize}[noitemsep]
		\item We introduce a new spatio-temporally annotated dataset, the CADP dataset, for accident forecasting using traffic camera views.
		Our dataset provides a novel view for traffic accident learning, and we hope to contribute to the enhancement of research on driving education as well as road safety.
		\item We apply state-of-the-art object detection models such as Faster R-CNN and accident forecasting models to our dataset and show their results.
		\item We exploit the contextual information around the object bounding box and test the impact of Context Mining and Augmented Context Mining within Faster R-CNN to improve the detection of small objects such as person and improve the Faster R-CNN baseline scores.
	\end{itemize}
	\section{Related Work}\vspace{-1mm}
	\noindent\textbf{Dataset for Car Modelling and Accidents}~With the development of the concepts of smart cities and autonomous driving, recent works target concerning traffic safety monitoring using computer vision techniques.
	\cite{Datondji2016} provides information about relevant datasets for traffic monitoring at road intersections:~the MIT dataset for traffic camera events~(a 19-min video), NGSIM dataset for road traffic modeling, CBSR dataset for single views at complex intersections, CVRR dataset which simulated videos generated for traffic modeling, QMUL dataset that contains recording at a busy intersection and KIT dataset which consists of videos with fog, rain and snow to model traffic car behaviour near intersections.
	\cite{HeXin2017} performs experimentation's using Faster R-CNN~\cite{Ren2015} to show the detection performance on the INRIA dataset.
	For the traffic accident videos, a recent UCF-Crimes dataset~\cite{Sultani2018} has 13 real-world anomalies such as Abuse, Accidents, Shooting and is focused on understanding of violent scenes in video.
	Dashcam Accident Dataset~(DAD)~\cite{Chan2016} uses Dashboard Camera captured videos to perform accident forecasting with 2.4 hours of video data.
	We believe that both Dashboard camera views and Traffic camera views could provide critical information for predicting accidents.
	However, traffic cameras give an overview of the complete road and thus will be able to track more vehicles as compared to dashboard camera views. 
	
	\noindent\textbf{Object Detection}~In recent years, object detection task gained pace and ~\cite{Girshick2014,Girshick2015,Ren2015,He2017,Liu2016} utilize the strength of deep learning~\cite{LeCun2015} in common benchmarks such as PASCAL VOC~\cite{pascal-voc-2012} and Microsoft COCO~\cite{Lin2014}. 
	R-CNN~\cite{Girshick2014} uses a region proposal algorithm as a pre-processing step prior to CNN architecture feature extraction. 
	These proposals are generated using Edge Boxes~\cite{Zitnick2014} or Selective Search~\cite{Uijlings13} and are independent of CNN.
	SPP-Net~\cite{He2015} were proposed to improve the R-CNN speed by sharing computation.
	Fast R-CNN~\cite{Girshick2015} reduces the run time exposing the region proposal computation as bottleneck whereas Faster R-CNN implements region proposal mechanism using CNN and thus integrating region proposal as part of the CNN training and prediction~\cite{Ren2015}.
	Mask R-CNN~\cite{He2017}, Single-Shot Detector~(SSD)~\cite{Liu2016} and FPN~\cite{Lin2017} combine multiple feature maps with different resolutions to naturally handle multiple object sizes.
	
	\noindent\textbf{Pedestrian Detection} predicts information about the pedestrian position based on the detection in current frame.
	\cite{Dollar2012} provides a comprehensive overview and arguments to replace continuous detection by pedestrian tracking and thus achieve real-time performance for pedestrian detection.
	\cite{Benenson2014} shows adding extra features, flow information and context information are complementary additions resulting in significant gains over other strong detectors. 
	\cite{Wang2018} uses body-part semantics and contextual information.
	\cite{Li2017} proposes a Haar-like cascade classifier design for fast pedestrian detection. 	\cite{Kong2018} proposes an extension to Faster R-CNN using contextual information with multi-level features to detect pedestrians in cluttered background obtaining embedding pooling information from a larger area around original area of interest.
	
	\noindent\textbf{Accident Detection and Forecasting}~In recent years, there have been a few works focusing on the use of cameras for accident forecasting. 
	For example,~\cite{Chan2016} uses Dashboard Cameras for accident forecasting. We believe that there is a strong requirement for those datasets to improve the reaction time of autonomous vehicles such as self-driving cars, and help the road surveillance.
	
	\section{Car Accidents Dataset}\vspace{-1mm}
	
	%\subsection{Dataset Statistics}\vspace{-1mm}
	Statistics of our dataset can be found in Table~\ref{tbl1} and Figure~\ref{fig2}.
	Data collection and annotation process are described in Appendix~\ref{appendix:A1}.
	\begin{figure*}[h]
	\centering
	    \subfigure[Number of objects by categories]{\includegraphics[clip,width=0.66\columnwidth]{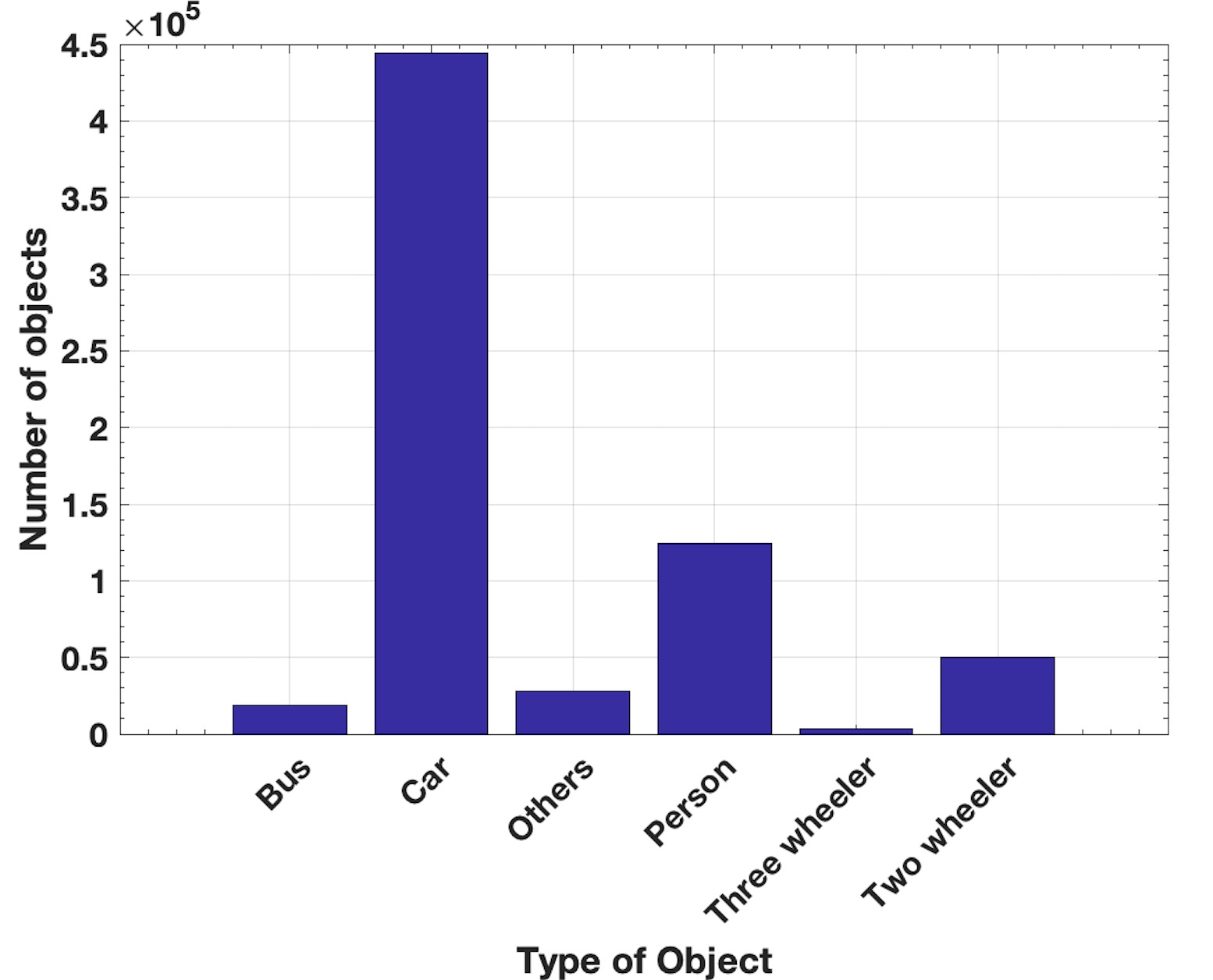}}
	    \subfigure[Number of usual/unusual tracks]{	\includegraphics[clip,width=0.66\columnwidth]{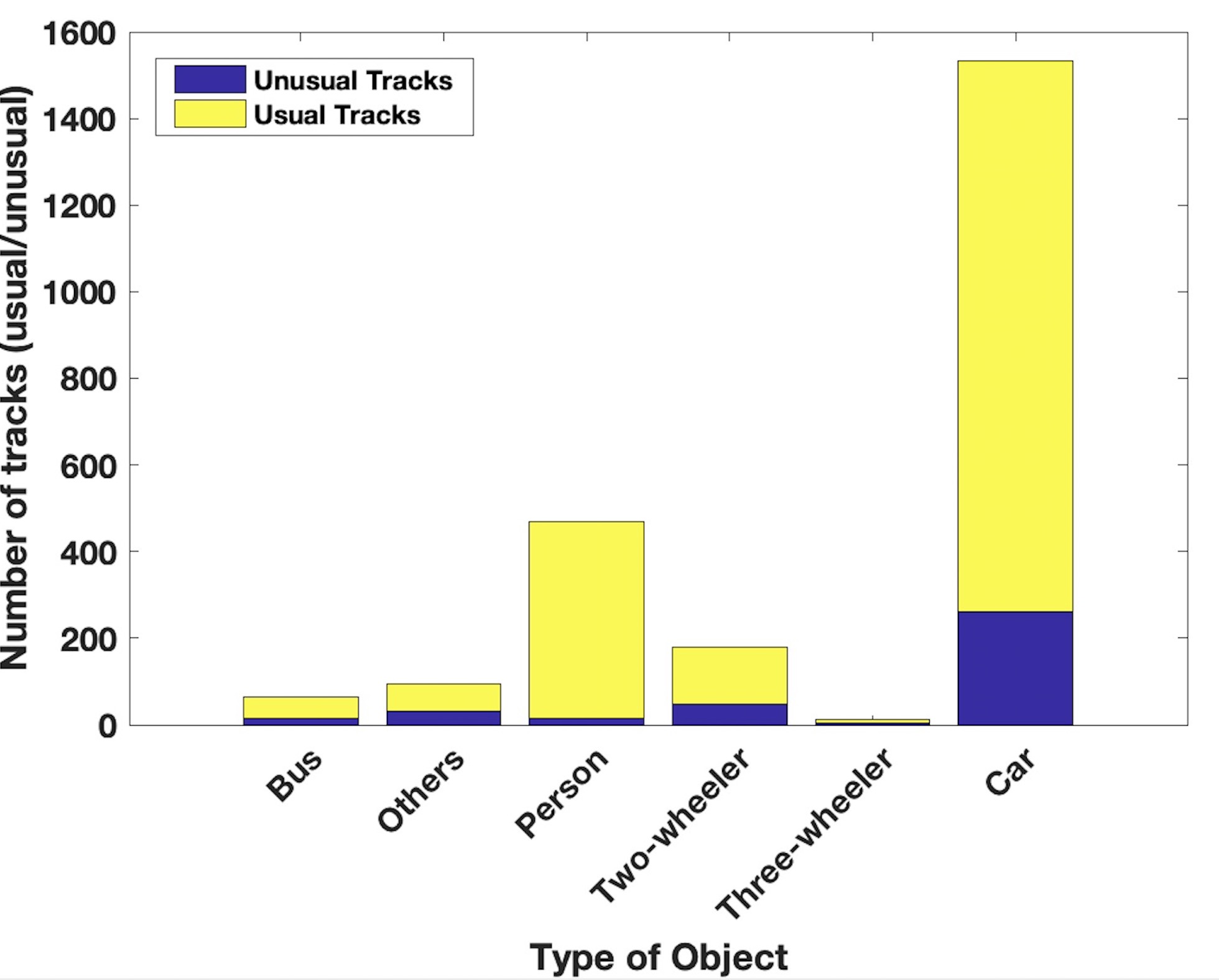} }
	    \subfigure[Number of objects by sizes]{	\includegraphics[clip,width=0.6\columnwidth]{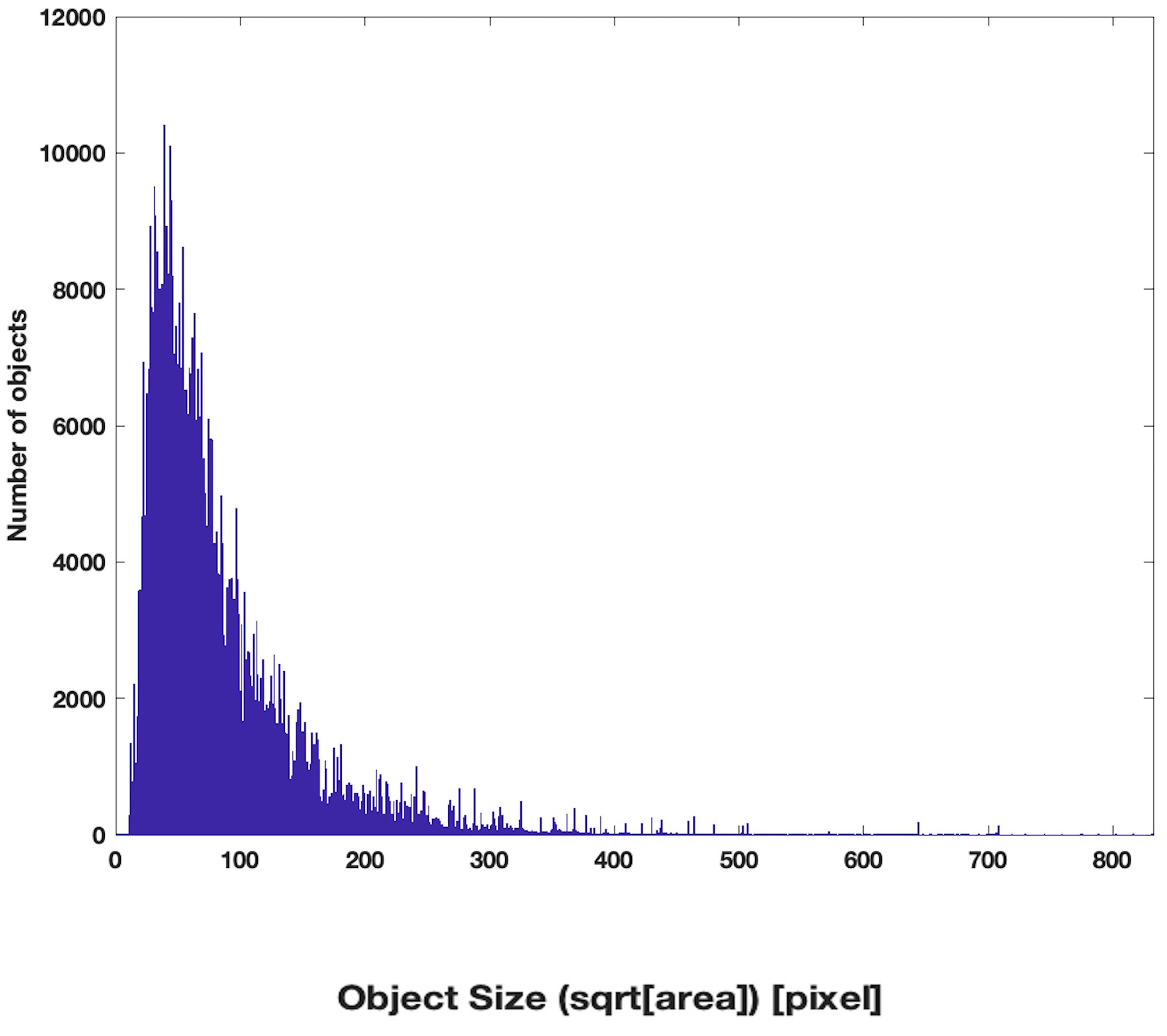}}\vspace{-2mm}
		\caption{The statistics of the CADP dataset.}
		\label{fig2}
	\end{figure*}
		\begin{table*}[h]
		\caption{Comparison between our dataset and related datasets.
			\textbf{T}: temporal annotation;
			\textbf{S}: spatial annotation (e.~g.~bounding boxes or pixel-level annotations);
			\textbf{A}: traffic accidents;
			\textbf{C}: videos were captured from a CCTV footage.
			The ``\# positives" refer to the number of videos which contain an accident.
			This statistics is computed from video-level labels~(no accident/accident).
			Our dataset is not the largest in terms of the number of hours, but is the largest in terms of number of accidents~(positive events).}
		\label{tbl1}
		\centering
		\begin{threeparttable}
		\begin{tabular}{lcccccccc}
			
			Dataset name    & \# videos                            & \# positives                      & Total duration                            & Avg. \# frames              & T & S & A & C \\\hline\hline
			UCF-Crimes~\cite{Sultani2018}      & \textbf{1900}\tnote{*}                                 & 151                               & \textbf{128 hours}\tnote{*}                                 & \textbf{7247}\tnote{*}                           & \cmark               & \xmark             & \cmark                 & \cmark            \\
			DAD~\cite{Chan2016}            & 1730                                 & 620                              & 2.4 hours                                 & 100                            & \cmark               & \cmark             & \cmark                 & \xmark            \\\hline
			\textbf{Ours}   & 1416 & {\textbf{1416}} & 5.2 hours & 366                   & \textbf{\cmark}      & \textbf{\cmark}    & \textbf{\cmark}        & \textbf{\cmark} \\\hline 
		\end{tabular}
		\begin{tablenotes}\footnotesize
        \item[*] These numbers from UCF-Crimes dataset are of 13 categories of crimes~(not only for traffic accidents).
        \end{tablenotes}
		\end{threeparttable}
	\end{table*}
	Some key characteristics of our dataset are as follows:
	\begin{itemize}[noitemsep]
		\item \textbf{Object size}:~As shown in Figure~\ref{fig2}(c), a major portion of the CADP dataset is occupied by small objects.
		Accurate detection of small objects has been a challenge in surveillance videos for a long time.
		The CADP dataset provides additional samples for these objects from traffic CCTV footage.
		\item \textbf{Video length}:~The average length of the videos in the CADP dataset is 366 frames per video, which is 3.66x longer than the dataset from ~\cite{Chan2016}.
		The longest video has 554 frames.
		The UCF-Crimes~\cite{Sultani2018} also has a category for road incidents with long videos, but only temporal annotations are provided.
		The CADP dataset provides a set of videos with full spatio-temporal annotations.
		\item \textbf{Number of positive videos}~(1416 videos) in our dataset for only traffic accidents is much larger than that in UCF-Crimes~(151 videos of road accidents) and DAD~(about 600 videos).
		Note that, in CADP, there are videos with more than one accident.
		Our dataset is devoted to traffic accidents~(positive events), and we did not collect videos of negative events.
		Negative segments can be critical for learning, but the presence of negative events can be found easily in other datasets such as DETRAC~\cite{lyu2017ua}.
		\item \textbf{Time to first accident} is the duration from time 0 in the video to the onset of the first accident.
		In the fully annotated subset of 205 videos in CADP dataset, this measure is 3.69 seconds in average.
		Compared to DAD~\cite{Chan2016}~(4.50 seconds), CADP has a shorter time-to-first-accident.
		This characteristic can affect the design of experimentation for accident forecasting.
		\item \textbf{Real-world data}:~CADP contains videos collected from YouTube which are captured under various camera types and qualities, weather conditions~(see Figure~\ref{fig1}) and edited/resampled videos.
	\end{itemize}
	
	\begin{figure*}[t]
		\centering
		\subfigure[Our workflow to solve the problem of accident forecasting.]{\includegraphics[clip, width=2.0\columnwidth]{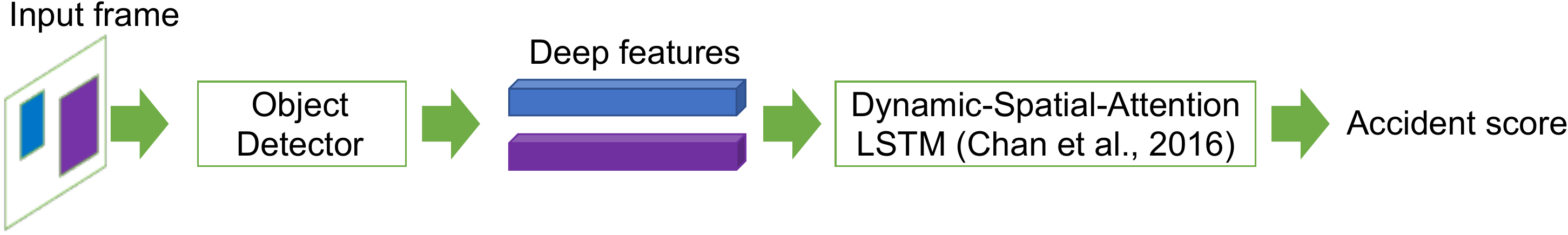}}\hfill
		\subfigure[Original Faster R-CNN architecture.]{\includegraphics[clip,width=0.8\columnwidth]{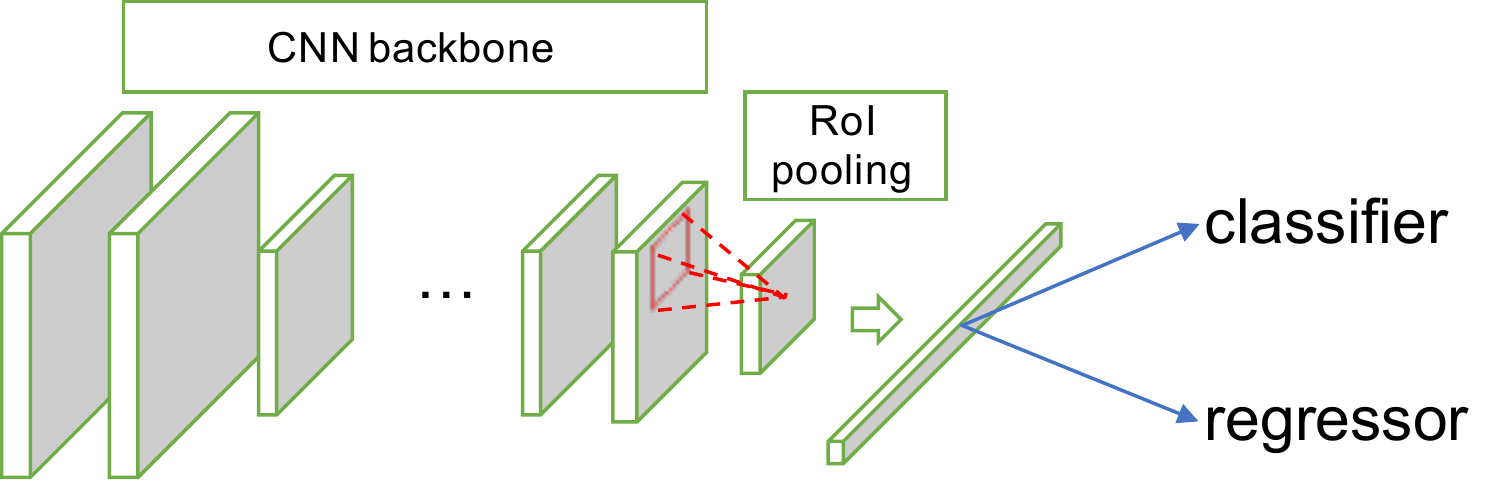}}
		\subfigure[Our proposed RoI pooling layers.]{\includegraphics[clip,width=0.8\columnwidth]{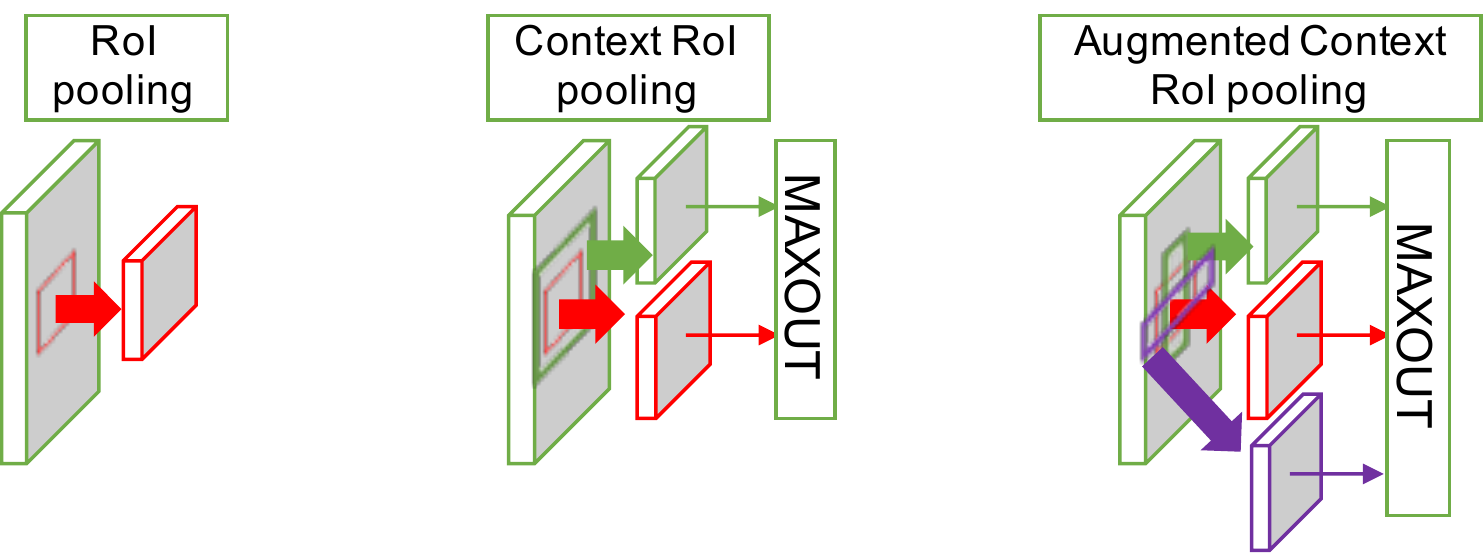}}
				\caption{Improved Faster R-CNN with Augmented Context Mining and a system for Accident Forecasting.}
		\label{fig4}
	\end{figure*}

	\section{Improved Faster R-CNN and Forecasting}\vspace{-2mm}
	\subsection{Improved Faster R-CNN for Object Detection}\vspace{-1mm}
	\noindent\textbf{Faster R-CNN}~\cite{Ren2015} is a deep learning architecture for object detection in still images.
	%It has been successfully applied to object detection in well-known benchmarks such as PASCAL VOC 2007/2012~\cite{pascal-voc-2012} and Microsoft COCO~\cite{Lin2014}, and recently in pedestrian detection domain~\cite{Zhang2016,Ren2018}.
	Like its preceders~\cite{Girshick2014,Girshick2015}, it extracts deep features of each proposal regions using a deep learning backbone such as ResNet-50.
	However, Faster R-CNN is an end-to-end architecture, because the proposal generation step is done using an internal proposal generation mechanism, the Region Proposal Network (RPN), which reduces the need for dependence on external proposal algorithm such as Selective Search or Edge Boxes, with a sliding window fashion.
	An important designing aspect of Faster R-CNN is its two-stage design:~after features are extracted for proposals, they are classified and regressed to match the anchor boxes.
	Learning in Faster R-CNN is done with objectives for bounding box regression and classification as follows:~$\mathcal{L}_{reg}=\sum_{i}\mbox{smooth}_{L1}(t_i-v_i)~,
	\mathcal{L}_{cls}=\sum_{i}-\log p_u~$.
	%\begin{eqnarray}
	%\mathcal{L}_{reg}&=&\sum_{i}\mbox{smooth}_{L1}(t_i-v_i)~,\\
	%\mathcal{L}_{cls}&=&\sum_{i}-\log p_u~,
	%\end{eqnarray}
	where $u$ and $v$ are the true class and target bounding box for a groundtruth anchor, $p$ and $t$ are the predicted probability of class $u$ and predicted bounding box.
	The $\mbox{smooth}_{L1}$ loss function is defined as in~\cite{Girshick2015}.
		
	\noindent\textbf{Implementation details}~We rescale the image to 600 pixels size to smallest size of the image as well as use 3 sizes for the anchor boxes 128$^{2}$, 256$^{2}$, and  512$^{2}$  pixels. Further, the aspect ratios of the anchor boxes is fixed at 1:1, 2:1 and 1:2 pixels as in Faster R-CNN paper~\cite{Girshick2015}.
	
	\noindent\textbf{Training procedure}~The multi-task objective for learning Faster R-CNNs is $\mathcal{L}=\mathcal{L}_{cls}+\lambda\mathcal{L}_{reg}$.
	%\begin{equation}
	%\mathcal{L}=\mathcal{L}_{cls}+\lambda\mathcal{L}_{reg}~.
	%\end{equation}
	For negative mining, we use the standard approach:~after the predicted boxes are filtered using non-maximum suppression (NMS) at the overlap threshold 0.7, the RoIs which have confidences in the range [0.1,0.5) are considered as ``hard negative", and the RoIs which have confidence larger than 0.5 are considered as ``positive".
	Finally, assuming that we need 32 candidates to contribute to the final loss, we randomly select positive RoIs first to fill at least 16 positions, then we randomly select from the negative RoIs to fill all 32 positions. 
	Only these candidates contribute to the final loss.
	For data augmentation, we use horizontal and vertical flips.
	
	\subsection{Context Mining}\vspace{-1mm}
	\noindent\textbf{Context Mining}~As noticed from Figure~\ref{fig2}(c), our dataset consists of objects which are small objects ($<$100 pixels) in majority.
	Moreover, from preliminary results on CADP using ResNet-50 backbone Faster R-CNN, we found that there is a significant degradation of accuracy~(mAP@0.5) for "Person"~(pedestrian) category.
	We argue that, the reason is because, when captured from CCTV traffic camera footage, a person occupies smaller areas than other vehicle categories in our dataset.
	Therefore, the bounding boxes of the pedestrians often contains fewer pixels than other objects.
	To this end, we propose to mine the context information around the small objects in CADP dataset, by extracting the context information in the RoI pooling layer~\cite{Girshick2015}.
	Given the region of interest of a small object $\mathbf{x}$, a context region $\mathbf{c}$ in common sense~(Figure~\ref{fig3}(a)) contains $\mathbf{x}$.
	By extending the context regions, more information is involved into the deep features.
	Let $\mathcal{C}=\{\mathbf{c}_i\}_{i=1}^n$ be the pooled contextual features, we choose the best responses from them by using Maxout networks~\cite{Goodfellow2013}. 
	By applying the dropout process to only linear parts of the signals, the Maxout network is considered to have more generalization ability than traditional Dropout approach.
	The Maxout operator is applied to $\mathcal{C}\cap \{\mathbf{x}\}$ to obtain final pooled feature $\mathbf{f} = \mbox{Maxout}(\mathcal{C}\cap \{\mathbf{x}\})$.
	%\begin{equation}
	%\mathbf{f} = \mbox{Maxout}(\mathcal{C}\cap \{\mathbf{x}\})~.
	%\end{equation}
	\noindent\textbf{Augmented Context Mining}~
	The bounding box annotations for small objects can be inaccurate due to human errors~(because the object size is too small and difficult for humans to draw a tight bounding box, annotators often draw a larger box or a box which truncates a part of the body).
	Furthermore, by enlarging the boxes, due to occlusion, the context may involve a different object into the box.
	Thus, to address these concerns, we also consider a different context mining, the Augmented Context Mining~(ACM) to fully exploit all possible patterns of context around the small person boxes.
	Rather than gradually extending the small regions to obtain the contexts, we narrow down and extend the small boxes in both horizontal and vertical directions.
	Given a step stride $s$ and the number of horizontal and vertical steps $m,n\in\{0, \pm 1,\pm 2,\ldots\}$, an \textit{augmented context} $\mathbf{a}=\mathbf{x}_{m,n}$ is defined by extending (when $m,n>0$) or narrowing down~(when $m,n<0$) $\mathbf{x}$.
	In the Results section, we compare the performance of these two mining strategies.
	
	\noindent\textbf{Implementation details}~To control the effects of CM/ACM on small objects, we introduce constraints based on the area ratio of the bounding box and the image.
	Given a bounding box with area $B$ and image with area $I$, and a threshold $\alpha\in [0,1])$.
	The context mining will be applied to a region if and only if $B\le \alpha S$.
	We choose $\alpha=0.01$ in our experiments.
	
	\subsection{Accident Forecasting}\vspace{-1mm}
	Our framework for Accident Forecasting can be found in Figure~\ref{fig4}(c).
	First, we extracted the features from the last fc layer~(2048D) in Faster R-CNN.
	The features are then fed into the Dynamic-Spatial-Attention LSTM~(DSA-LSTM)~\cite{Chan2016} to output accident scores over time.
	DSA-LSTM is built upon the famous Soft-Attention LSTM~\cite{Xu2015}.
	However, instead of applying spatial attention to regular grid, DSA-LSTM distributes the attentional weights to spatial objects detected by a state-of-the-art detector~\cite{Ren2015}.
	Furthermore, DSA-LSTM applies to ``sequences" of frames dynamically~(when Soft-Attention LSTM applies to a single frame for caption generation).
	The full-frame features are also exploited and exponential loss is applied for training with positive sequences.
	In our view, the exponential loss fits the nature of traffic accidents in CADP because accidents often happen \textit{suddenly} and the damages grow exponentially in a short time.
	The exponential loss for positive events can be formulated as follows:~$\mathcal{L}_p\left(\left\{\mathbf{a}\right\}\right)=\sum_{t}-e^{-\max (0,y-t)}\log (a_t)$,
	%\begin{equation}
	%\mathcal{L}_p\left(\left\{\mathbf{a}\right\}\right)=\sum_{t}-e^{-\max (0,y-t)}\log (a_t)~,
	%\end{equation}
	where $\mathbf{a}$ is the attended object, $y$ is the time the accident happens, and $a_t$ is the accident probability of $\mathbf{a}$ at time $t$.
	For the negative sequences~(no accidents), we used cross-entropy loss:~$\mathcal{L}_n\left(\left\{\mathbf{a}\right\}\right)=\sum_{t}-\log (a_t)$.
	%\begin{equation}
	%\mathcal{L}_n\left(\left\{\mathbf{a}\right\}\right)=\sum_{t}-\log (a_t)~.
	%\end{equation}
	\begin{figure}[tb]
		\centering
		\subfigure[]{\includegraphics[width=0.45\columnwidth]{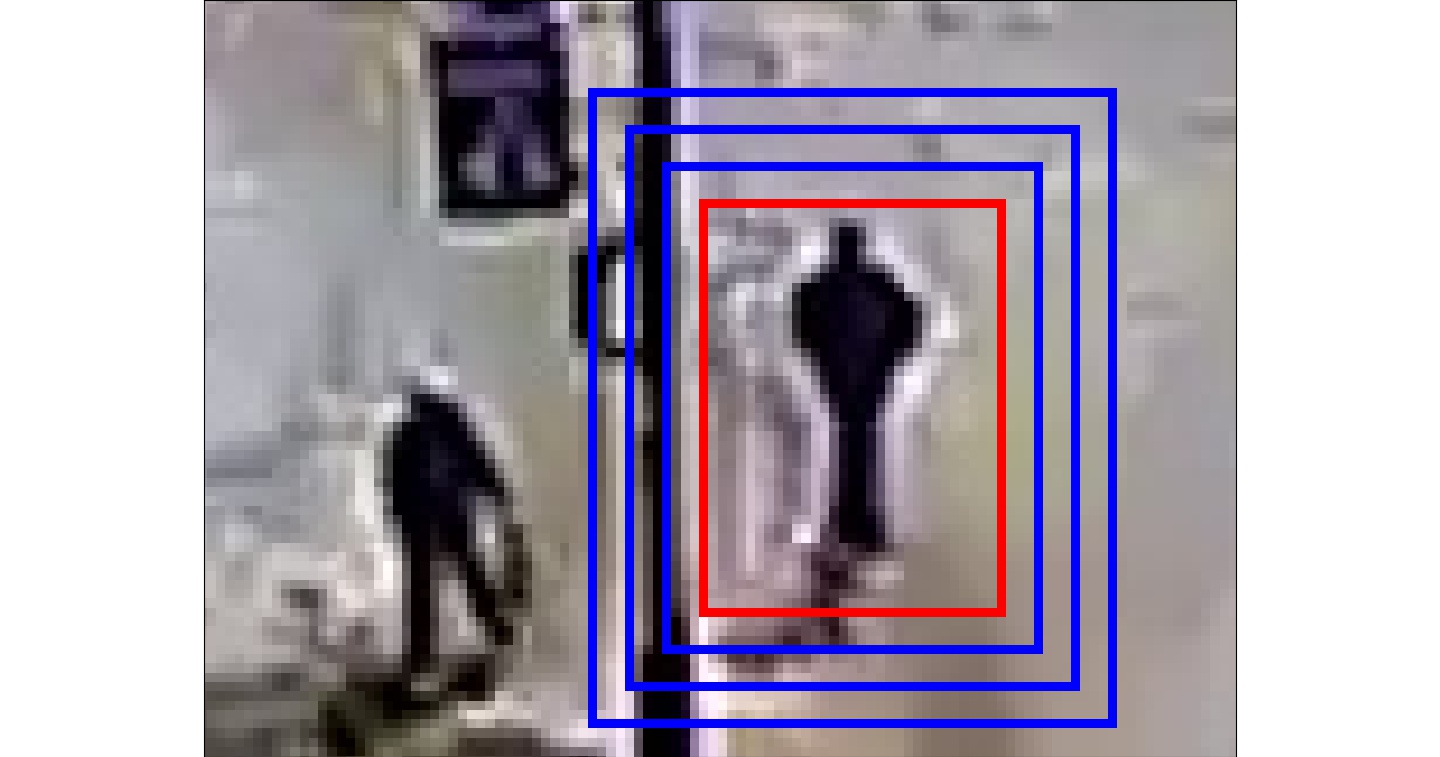}}
		\subfigure[]{\includegraphics[width=0.45\columnwidth]{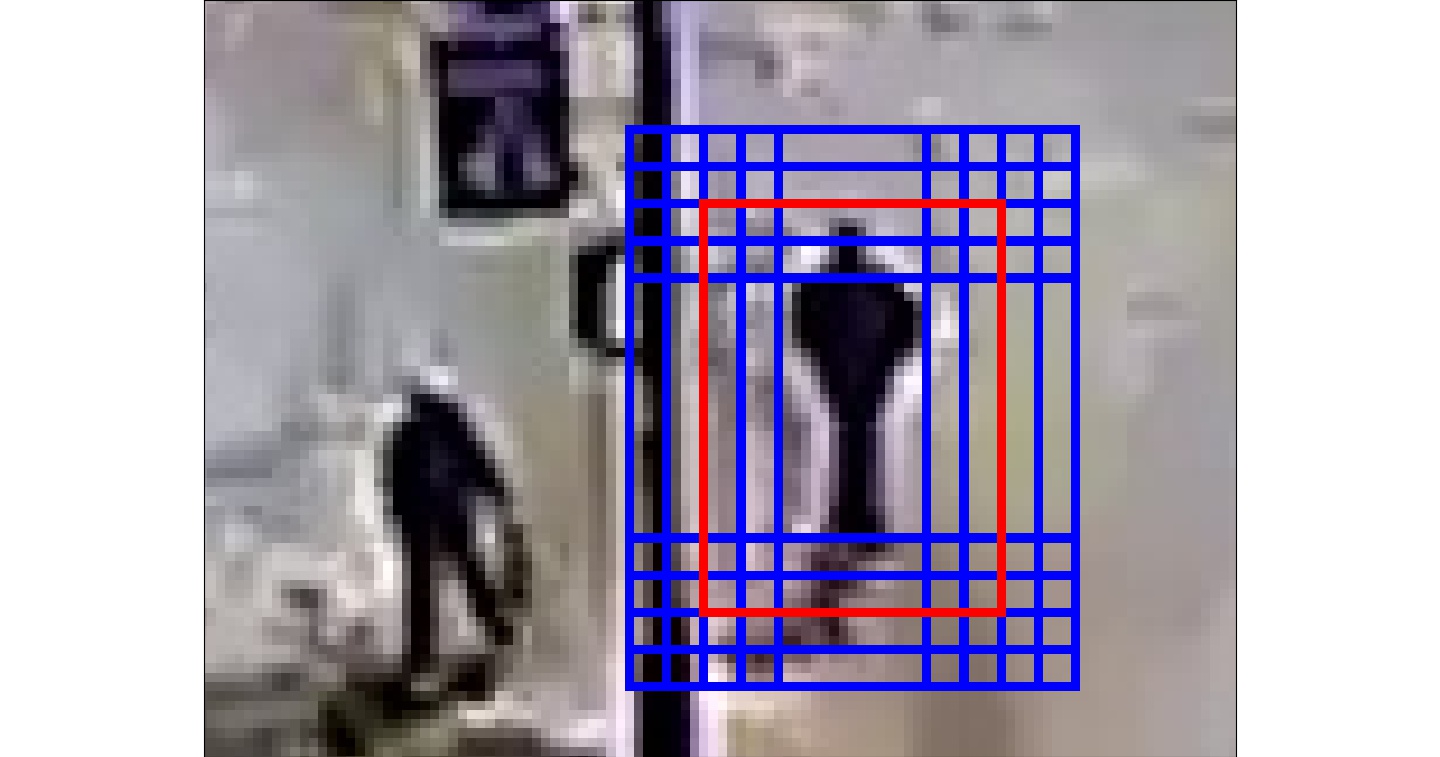}}
		\caption{Contextual patterns.
		\textcolor{red}{Red} bounding boxes indicate the original labels provided by human annotators.
		\textcolor{blue}{Blue} bounding boxes indicate contexts mined by our algorithms.
			(a)~The \textit{contextual} bounding box is created by extending the small object region by $s=2$ pixels in horizontal and vertical directions.
			The information of the original labels~(red box) is preserved.;
			(b)~The \textit{augmented contextual} bounding boxes are created by extending or narrowing down the horizontal and vertical sides by $s=2$ pixels.
			Although there are more bounding boxes with diverse information, there are also many negative bounding boxes~(boxes inside the red frame) which do not preserve information of original label.}
		\label{fig5}
	\end{figure}
	
	\noindent\textbf{Exhaustive negative mining}~Negative examples are often critical for learning in various situations.
	However, CADP does not provide explicit negatives.
	A potential source to mine these examples is existing datasets such as the DETRAC dataset~\cite{lyu2017ua}.
	In this work, we exhaustively mine the negatives from positive sequences.
	Given an accident happens at time $t$, we mine a positive segment with length 100 frames from time $t-90$ to time $t+10$.
	We randomly mine a segment with length 100 frames which does not overlap with the positive event.
	Because our videos are longer than 100 frames in average, this mining scheme was possible.
	However, many accidents happen between the first 100 frames and the Time-to-First-Accident in our test set is only 3.69 seconds, therefore sampling from $t-90$ may not be possible.
	To exhaustively mine the negative segments, we append the dummy frames before time 0 to have 90 frames.

	\section{Results}
	\subsection{Experimental setup}
	\begin{table*}[h]
		\centering
		\caption{Comparisons between state-of-the-art methods in our dataset.
		We choose SSD and Faster R-CNN because they are the popular choices for object detection in surveillance video literature.}\vspace{-1mm}
		\label{tbl3}
		\begin{tabular}{cccccccc}
			\textbf{Method} & \textbf{All} & \textbf{Person} & \textbf{Car} & \textbf{Bus} & \textbf{Two-wheeler} & \textbf{Three-wheeler} & \textbf{Others} \\\hline\hline
			SSD~\cite{Liu2016}             & 64.70        & -               & -            & -            & -                    & -                      & -               \\
			Faster R-CNN~\cite{Ren2015}    & 84.39          & 52.26           & 89.39          & 97.27          & 77.56                & 98.88                  & 91.00                     
		\end{tabular}
	\end{table*}
    \begin{table*}[h]
	\centering
	\caption{Cross-validation results for Faster R-CNN.
		We used mAP@0.5~(0.5 is IOU score) as the measure.
	Except the ``Person" category, Faster R-CNN performs stably  across all categories of vehicles.}
	\label{tbl2}
	\begin{tabular}{cccccccc}
		\textbf{Fold} & \textbf{All}   & \textbf{Person} & \textbf{Car}   & \textbf{Bus}   & \textbf{Two-wheeler} & \textbf{Three-wheeler} & \textbf{Others} \\\hline\hline
		1             & 82.32          & 36.89           & 81.04          & 97.24          & 76.00                & 98.71                  & 94.58           \\
		2             & 84.39          & 52.26           & 89.39          & 97.27          & 77.56                & 98.88                  & 91.00           \\
		3             & 84.33         & 47.22           & 85.40          & 98.57          & 82.32               & 98.30                  & 94.58           \\\hline
		\textbf{Mean} & \textbf{83.68} & \textbf{45.46}  & \textbf{85.28} & \textbf{97.69} & \textbf{78.63}       & \textbf{98.63}         & \textbf{93.39} 
	\end{tabular}
    \end{table*}
    
    \noindent\textbf{Cross-validation}~%is critical to assure the stability of the models.
	We sample a \textit{trainval} set of 103 videos for training of object detectors and accident forecasters.
	The 102 remaining videos have been used to test the forecasters. Our choice was contingent on creating a robust model which we wanted to test on enough samples and thus split with a 50:50 ratio (train and test set) where each set has similar set statistics in terms of number of objects. For object detection, from the frames of the 103 videos in \textit{trainval} set, we sample randomly three folds (train/test split) to compute the accuracy.
	After the cross-validation of object detectors in \textit{trainval} set, we select the best performers as the feature extractor for training the accident forecaster~(see Figure~\ref{fig4}(c)).
	
	\noindent\textbf{Implementation details}~Our system is implemented using the Tensorflow framework\footnote{\url{https://www.tensorflow.org/}}.
	%For object detection with Faster R-CNN, all images are resized to have the same height (600 pixels). % Already mentioned in previous section.
	During testing, we improve performance by detecting objects with different scales of images~(multi-scale testing).
	For SSD, we use the implementation of~\cite{Liu2016}.
	We fine-tune all object detectors in the CADP \textit{trainval} set until convergence.
	For accident forecasting, we follow the details described in the previous section.
	The initial learning rate for Faster R-CNN was $10^{-5}$ and the Adam optimizer was used.
	
	\noindent\textbf{Evaluation measures}~For object detection, we use mean Average Precision at IoU=0.5~(mAP@0.5) ~\cite{pascal-voc-2012} to assess the accuracy of the detectors.
	For accident forecasting, we follow~\cite{Chan2016} and use Time-to-Accident~(ToA) and recall, precision and average precision~(AP).
	%Given the number of true-positives~(TP), false-positives~(FP) and false-negatives~(FN), $Precision=\frac{TP}{TP+FP}$, and $Recall=\frac{TP}{TP+FN}$.
	To compute the AP, we sample various thresholds and compute ToA, recall and precision at each operating point.
	AP and mean ToA are computed from these data.
	
	\subsection{Object Detection}\vspace{-1mm}
	\begin{table}[h]
		\centering
		\caption{Ablation study on different object detectors.
		$s$ is the step stride to extend or narrow down the width/height of a context, $n_c$ is the number of contexts in Context Mining, and $m,n$ are the parameters of ACM.}
		\label{tbl4}
		\begin{tabular}{lccc}
			\textbf{Method}            & \textbf{Parameter} & \textbf{$s$} & \textbf{mAP@0.5} \\\hline\hline
			Faster R-CNN& -     & - &  84.33       \\\hline
			\multirow{8}{*}{Context Mining} & $n_c=2$ & \multirow{4}{*}{2} &     70.52    \\
			& $n_c=4$ &  &     81.00    \\
			& $n_c=8$ &  &     90.49  \\
			& $n_c=16$ & & 92.83\\\cline{2-4}
			& $n_c=2$ & \multirow{4}{*}{4} &     77.43    \\
			& $n_c=4$ &  &    89.04    \\
			& $n_c=8$ &  &     92.59  \\
			& $n_c=16$ & &\textbf{92.84}\\\hline
			\multirow{1}{*}{Augmented CM} & $m=n=8$ & \multirow{1}{*}{4}  &    90.53     \vspace{-1mm}
		\end{tabular}
	\end{table}

\begin{table}[h]
	\centering
	\caption{Person detection results of the best methods.}\vspace{-1mm}
	\label{tbl6}
	\begin{tabular}{lcc}
		\textbf{Method}            & \textbf{mAP@0.5} & \textbf{Improvement}\\\hline\hline
		Faster R-CNN &  47.22 & -      \\
		Context Mining & 93.67 &     \textbf{+46.45}    \\ 
		Augmented CM & 92.44 & +45.22  \vspace{-1mm}
	\end{tabular}
\end{table}
	\begin{table}[h]
		\centering
		\caption{Performance comparison between different accident forecasters.
		ToA@0.8 is the ToA when Recall is 80.0\%.
		The results are obtained after training each models for 40 epochs like in~\cite{Chan2016}.}
		\label{tbl5}
		\begin{tabular}{lccc}
			\textbf{Method} & \textbf{AP} & \textbf{mToA} & \textbf{ToA@0.8} \\\hline\hline
			DSA~\cite{Chan2016}&     \textbf{47.36}        &         1.359                  &       1.798               \\
			ACM+DSA&        47.09     &            1.457               &  2.104   \\
			CM+DSA&        47.25  &     \textbf{1.684}      &     \textbf{3.078}         
		\end{tabular}
	\end{table}
	\noindent\textbf{Baselines:~SSD~vs.~Faster R-CNN}~The comparison between SSD and Faster R-CNN can be found in Table~\ref{tbl3}.
	Interestingly, we observed a large gap between the mAP@0.5 of SSD and Faster R-CNN~(approx.~19.69\%).
	From the observation about the performance of these two detectors, we choose Faster R-CNN as the baseline for further experimentation.
	The performance of Faster R-CNN over three sampled folds are reported in Table~\ref{tbl2}.
	We can observe stable performances of this detector in CADP \textit{trainval} set.
	However, we can also observe that the performances degrade and become unstable in the ``Person" category.
	
	\noindent\textbf{Context Mining}~
	We choose the third fold to perform ablation study on hyper-parameter of Context Mining~(CM) and Augmented Context Mining~(ACM).
	The results are shown in Table~\ref{tbl4}.
	With appropriate hyper-parameters~($n_c=16, s=4$ for CM), CM and ACM significantly outperform the baseline~(+8.51\% for CM and +6.20\% for ACM).
	For pedestrian detection, results in Table~\ref{tbl6} indicate significant improvements of CM and ACM over Faster R-CNN.
	Between CM and ACM, CM outperforms ACM by about two points in terms of mAP@0.5.
	It implies that mining by small number of contexts and by extending the original regions gradually can lead to a better performance. 
	
	\noindent\textbf{Runtime analysis}~Increasing the number of contexts results in an increase of the inference time:~for Faster R-CNN, it takes 0.56 seconds for inference of a single image, while CM~($n_c=16,s=4$) takes 1.04 seconds and ACM~($m=n=8,s=4$) takes 5.81 seconds, with single GPU.
	Thus, mining in a large space of contexts requires time and resources.\vspace{-1mm}
	\subsection{Accident forecasting}\vspace{-1mm}
	The results for accident forecasting using DSA-LSTM~\cite{Chan2016} can be found in Table~\ref{tbl5}. For a dataset with average Time-to-First-Accident is 3.84 seconds, we can issue warning prior to the accidents at 1.359 seconds with highest AP is 47.36\%. Moreover, when recall is 80\%, the ToA is 1.798 seconds.
	For Context Mining, the results are 1.684 seconds and 3.078 seconds, respectively.
	Using CM features leads to better forecasting results.

\section{Conclusion}\vspace{-2mm}
We introduced the Car Accident Detection and Prediction~(CADP) Dataset from CCTV Traffic Camera videos. 
A detailed account of the challenges faced in creation of the dataset such as data collection, access to traffic camera footage were tackled in the paper.
We presented the results of state-of-the-art object detection and accident forecasting models on our dataset.
We highlighted the strengths and weaknesses of these baseline models, and outperformed the initial results by adding context mining or augmented context mining.
We finally showed that augmented context mining does not improve the score obtained with a gradual context mining for object detection.
We demonstrate the final model for accident forecasting that can predict accidents about 2 seconds before they occur with 80\% recall.

{\small 
\bibliographystyle{ieee}
\bibliography{egbib}

\begin{thebibliography}{10}\itemsep=-1pt

\bibitem{Benenson2014}
R.~Benenson, M.~Omran, J.~Hosang, and B.~Schiele.
\newblock {Ten Years of Pedestrian Detection, What Have We Learned?}
\newblock In {\em ECCV 2014 Workshops}, pages 613--627, 2014.

\bibitem{Chan2016}
F.~H. Chan, Y.~T. Chen, Y.~Xiang, and M.~Sun.
\newblock {Anticipating accidents in dashcam videos}.
\newblock In {\em Asian Conference on Computer Vision}, 2016.

\bibitem{Datondji2016}
S.~R.~E. Datondji, Y.~Dupuis, P.~Subirats, and P.~Vasseur.
\newblock {A Survey of Vision-Based Traffic Monitoring of Road Intersections}.
\newblock {\em IEEE Transactions on Intelligent Transportation Systems},
  17(10):2681--2698, oct 2016.

\bibitem{Dollar2012}
P.~Dollar, C.~Wojek, B.~Schiele, and P.~Perona.
\newblock {Pedestrian Detection: An Evaluation of the State of the Art}.
\newblock {\em IEEE Transactions on Pattern Analysis and Machine Intelligence},
  34(4):743--761, apr 2012.

\bibitem{pascal-voc-2012}
M.~Everingham, L.~Van~Gool, C.~K.~I. Williams, J.~Winn, and A.~Zisserman.
\newblock The {PASCAL} {V}isual {O}bject {C}lasses {C}hallenge 2012 {(VOC2012)}
  {R}esults.
\newblock
  http://www.pascal-network.org/challenges/VOC/voc2012/workshop/index.html.

\bibitem{Girshick2015}
R.~Girshick.
\newblock {Fast R-CNN}.
\newblock In {\em IEEE International Conference on Computer Vision}, 2015.

\bibitem{Girshick2014}
R.~Girshick, J.~Donahue, T.~Darrell, and J.~Malik.
\newblock {Rich Feature Hierarchies for Accurate Object Detection and Semantic
  Segmentation}.
\newblock In {\em IEEE Computer Society Conference on Computer Vision and
  Pattern Recognition}, pages 580--587. IEEE, jun 2014.

\bibitem{Goodfellow2013}
I.~J. Goodfellow, D.~Warde-Farley, M.~Mirza, A.~Courville, and Y.~Bengio.
\newblock {Maxout Networks}.
\newblock In {\em International Conference on Machine Learning}, 2013.

\bibitem{He2017}
K.~He, G.~Gkioxari, P.~Dollar, and R.~Girshick.
\newblock {Mask R-CNN}.
\newblock In {\em IEEE International Conference on Computer Vision}, pages
  2980--2988. IEEE, oct 2017.

\bibitem{He2015}
K.~He, X.~Zhang, S.~Ren, and J.~Sun.
\newblock {Spatial Pyramid Pooling in Deep Convolutional Networks for Visual
  Recognition}.
\newblock {\em IEEE Transactions on Pattern Analysis and Machine Intelligence},
  2015.

\bibitem{HeXin2017}
X.~He and D.~Zeng.
\newblock {Real-time pedestrian warning system on highway using deep learning
  methods}.
\newblock In {\em International Symposium on Intelligent Signal Processing and
  Communication Systems (ISPACS)}, pages 701--706. IEEE, nov 2017.

\bibitem{Kong2018}
W.~Kong, N.~Li, T.~H. Li, and G.~Li.
\newblock {Deep Pedestrian Detection Using Contextual Information and
  Multi-level Features}.
\newblock In {\em International Conference on Multimedia Modeling (MMM)}, pages
  166--177, 2018.

\bibitem{LeCun2015}
Y.~A. LeCun, Y.~Bengio, and G.~E. Hinton.
\newblock {Deep learning}.
\newblock {\em Nature}, 2015.

\bibitem{Li2017}
J.~Li, X.~Liang, S.~Shen, T.~Xu, J.~Feng, and S.~Yan.
\newblock {Scale-aware Fast R-CNN for Pedestrian Detection}.
\newblock {\em IEEE Transactions on Multimedia}, pages 1--1, 2017.

\bibitem{Lin2017}
T.~Y. Lin, P.~Doll{\'{a}}r, R.~Girshick, K.~He, B.~Hariharan, and S.~Belongie.
\newblock {Feature pyramid networks for object detection}.
\newblock In {\em IEEE Computer Society Conference on Computer Vision and
  Pattern Recognition}, 2017.

\bibitem{Lin2014}
T.-Y. Lin, M.~Maire, S.~Belongie, J.~Hays, P.~Perona, D.~Ramanan,
  P.~Doll{\'{a}}r, and C.~L. Zitnick.
\newblock {Microsoft COCO: Common Objects in Context}.
\newblock In {\em European Conference on Computer Vision}, pages 740--755,
  2014.

\bibitem{Liu2016}
W.~Liu, D.~Anguelov, D.~Erhan, C.~Szegedy, S.~Reed, C.-Y. Fu, and A.~C. Berg.
\newblock {SSD: Single Shot MultiBox Detector}.
\newblock In {\em European Conference on Computer Vision}, pages 21--37, 2016.

\bibitem{lyu2017ua}
S.~Lyu, M.-C. Chang, D.~Du, L.~Wen, H.~Qi, Y.~Li, Y.~Wei, L.~Ke, T.~Hu,
  M.~Del~Coco, et~al.
\newblock Ua-detrac 2017: Report of avss2017 \& iwt4s challenge on advanced
  traffic monitoring.
\newblock In {\em Advanced Video and Signal Based Surveillance (AVSS), 2017
  14th IEEE International Conference on}, pages 1--7. IEEE, 2017.

\bibitem{Ren2015}
S.~Ren, K.~He, R.~Girshick, and J.~Sun.
\newblock Faster r-cnn: Towards real-time object detection with region proposal
  networks.
\newblock In {\em Advances in Neural Information Processing Systems 28}, pages
  91--99, 2015.

\bibitem{Sultani2018}
W.~Sultani, C.~Chen, and M.~Shah.
\newblock {Real-world Anomaly Detection in Surveillance Videos}.
\newblock In {\em IEEE Computer Society Conference on Computer Vision and
  Pattern Recognition}, 2018.

\bibitem{nationalreport1}
{The National Safety Council}.
\newblock The state of safety - a state-by-state report, 2017.

\bibitem{Uijlings13}
J.~Uijlings, K.~van~de Sande, T.~Gevers, and A.~Smeulders.
\newblock Selective search for object recognition.
\newblock {\em International Journal of Computer Vision}, 2013.

\bibitem{Vondrick2013}
C.~Vondrick, D.~Patterson, and D.~Ramanan.
\newblock {Efficiently Scaling up Crowdsourced Video Annotation}.
\newblock {\em International Journal of Computer Vision}, 101(1):184--204, jan
  2013.

\bibitem{Wang2018}
S.~Wang, J.~Cheng, H.~Liu, F.~Wang, and H.~Zhou.
\newblock {Pedestrian Detection via Body-part Semantic and Contextual
  Information with DNN}.
\newblock {\em IEEE Transactions on Multimedia}, pages 1--1, 2018.

\bibitem{Xu2015}
K.~Xu, J.~Ba, R.~Kiros, K.~Cho, A.~Courville, R.~Salakhudinov, R.~Zemel, and
  Y.~Bengio.
\newblock Show, attend and tell: Neural image caption generation with visual
  attention.
\newblock In {\em International Conference on Machine Learning}, pages
  2048--2057, 2015.

\bibitem{Zitnick2014}
C.~L. Zitnick and P.~Doll{\'{a}}r.
\newblock {Edge Boxes: Locating Object Proposals from Edges}.
\newblock In {\em European Conference on Computer Vision}, pages 391--405,
  2014.

\end{thebibliography}
}

\appendix

\section{Data collection and annotation}\label{appendix:A1}

	\begin{figure*}[tb]
		\centering
		\includegraphics[clip, width=1.8\columnwidth]{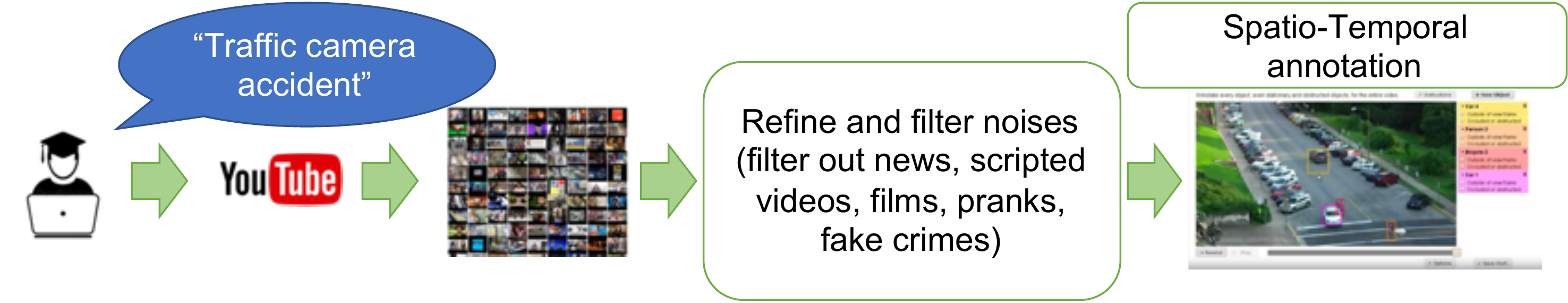}\vspace{-2mm}
		\caption{Data collection and annotation for traffic CCTV videos.}
		\label{fig3}
	\end{figure*} 
\subsection{Data collection}\vspace{-1mm}
	The major challenge in collecting data for traffic accidents is two-fold:~(i)~\textit{Abnormality}: because the accidents are rare, although there are live-streams from traffic cameras mounted on the corner of road intersections, this is infeasible to wait for an accident to happen; and~(ii)~\textit{Access}: access to traffic camera data is often limited. 
	Due to this challenge, the data of traffic accidents from fixed third-person views is often not available for public uses. To this end, in this work, we attempted to exploit an edge case, the traffic accidents captured from traffic camera views available on video sharing websites such as YouTube.
	The whole pipeline for data collection and annotation can be seen in Figure~\ref{fig3}.
	
	\noindent\textbf{Keyword search}~To collect the data for traffic accidents, we exploited the search engine and resources available in YouTube.
	We used keywords like "car accidents traffic camera" to search for relevant videos. This step returned 582 YouTube videos.
	
	\noindent\textbf{Refinement}~
	However, the collected videos from these queries contain many irrelevant items.
	To collect only relevant items, we employ three annotators to manually watch and report items as follows.
	All annotators are instructed to know that our objective is to collect only videos which \textit{contain at least one accident scene which is captured from a traffic CCTV footage}.
	The annotators then watched all collected videos one by one, and answered a survey about the videos.
%	~(see Figure~\ref{fig6}).
	Besides basic questions to identify whether the annotators want to download the videos based on explained objectives, there are three follow-up questions to filter noisy responses.
	The first question asks the annotators to justify their concrete reasons for downloading the videos.
	The second and third questions ask annotators about side aspects of the videos to discover inconsistency in their responses.
	Videos with inconsistent responses will be removed.
	
	\subsection{Annotations}
	After the refinement step, there are 230 videos that were found to be strongly relevant to our objectives. However, for each video, there is only a portion relevant to traffic CCTV footage.
	Therefore, we employed a two-stage annotation process to get these relevant segments:~first we asked human annotators to extract the starting and ending time-stamps for CCTV traffic camera segments from each videos, then we collected the segments and perform the spatio-temporal annotation using the VATIC tool~\cite{Vondrick2013}~(see Figure~\ref{fig3}).
	
	\noindent\textbf{Stage 1:~Temporal segmentation}~Most of the YouTube videos have a duration of several minutes but contain only several seconds with accidents from traffic CCTV footage. 
	Using the BeaverDam tool\footnote{\url{https://github.com/antingshen/BeaverDam}}, human annotators reported the starting and ending timestamps of each relevant segment.
	Based on the reported results, we extracted the frames of relevant segments using OpenCV\footnote{\url{https://opencv.org/}}.
	
	\noindent\textbf{Stage 2:~Dense Spatio-Temporal annotation}~After Stage 1, we have 1416 video segments of positive events.
	The total duration is 5.24 hours with an average number of frames of 366 frames per video~(see Table~\ref{tbl1}).
	About 80\% of videos have a length from 100 to 600 frames.
	From short videos (less than 600 frames), we choose 240 videos with HD quality to do dense spatio-temporal annotation.
	This stage involved four human annotators and has been done in about two months.
	From the 240 selected videos, the annotators identified 35 videos which are duplicated with one of the other videos.
	They are the videos with identical contents or cropped (resized) versions of another video.
	Finally, we have 205 videos with full annotations.
	The categories of objects are ``Person", ``Car"~(including minivans), ``Bus", ``Two-wheeler"~(including cyclists, motorbikes), ``Three-wheeler" and ``Others"~(objects which are not classified in other categories).
	About temporal annotations, human annotators were asked to mark when a collision between vehicles/pedestrians on the road happens, and when it ends.
	
	\section{Quality assurance}
	We hired two human annotators to do the spatio-temporal annotation using VATIC tool. During the trial time, we discovered that the quality of the submitted works by human annotators was low. Therefore, we extensively applied common practices to ensure the quality and confidence.
	\begin{itemize}
	    \item \textbf{Peer-review:}~First, we found three expert reviewers (validators) to hold the reviewing process. 
	    These experts agreed to work voluntarily (no payment). 
	    Their responsibility is to review the submitted works from human annotators and take appropriate actions: (i) recommend rejections to clearly bad quality works; (ii) recommend acceptances to clearly good works; and (iii) make further investigations on works lying around borderline. 
	    The three experts sometimes directly made modifications based on the guideline (which will be described later) to make works lying slightly under borderline meet the standard and then the works can get accepted later.
        \item \textbf{Annotation guideline:}~A common error we obtained during trial time is that there are misunderstanding between reviewers and annotators. 
        To resolve this problem, we made a 8-page annotation guideline to share the knowledge between expert reviewers and annotators. 
        The reviewers contributed to finish the guideline based on their experience and expertise about the process. 
        The definition of objects, annotation procedures and example of bad works are covered in the guideline. 
        These definitions and procedures were decided based on references (for example, definitions from dictionaries and literatures) and experiences of experts. 
        Then, we distributed the guideline to the annotators and asked them to read the guideline carefully. 
        The guideline is useful to reduce misunderstanding, and also helped to identify bad workers who did not pay attention to the guideline. 
        We asked reviewers to report such doubtful works.
        \item \textbf{Pretraining:}~the content of this work can be unfamiliar to the human annotators, especially to beginners. 
        The two human annotators had no experiences about spatio-temporal annotation before we asked. 
        Therefore, we spend a trial time with a number of small tasks to let them understand the work. 
        During this trial time, three experts provided feedback to grow annotators using chat and guidelines. 
        The trial time is also voluntarily with very low payment rate (almost 0 cent per hour, compared to payment rate during the official working time 15\$/hour).
	\end{itemize}
The 8-page annotation guideline and a different report which detailed the process are already publicly available in the Internet. 
Here we share several key insights we obtained from the work.
\begin{itemize}
    \item \textbf{Performance during trial time (pretraining) is quite low from the standard.} 
    After the first trial, one of our reviewers shouted that “we could not barely accept 1\% of your (the annotators’) works”.
    The communicator had to modify this sentence before sending to the annotators, although it can be honest. 
    Later, that sentence was even written into the guideline by the reviewer.
    \item \textbf{After performance meet the standard, the quality become better, and even the annotators start to provide feedback to the reviewers.} 
    Most of the feedback from annotators is about software problems, server is not accessible, etc.
    \item \textbf{Collaborative editing can lead to a mess.}
    If more than one annotator/reviewer modify a work at the same time, unsaved work of one can be disregarded by another one. 
    To resolve this problem, we first ask the annotators to finish their work and submit to the reviewers. 
    Then, the submitted works are divided into equal episodes without overlaps which were assigned to reviewers in a one-to-one mapping fashion. 
    Reviewers then returned bad works to the annotators for modifications with the list of errors. 
    Good works were then immediately saved and those cases are closed. This process has been repeated until all works were finished.
    \end{itemize}
Although better and more secure solutions such as Amazon Mechanical Turk (AMT) are available, due to the volume of works is not too large and we wanted to save money whenever volunteers are available, we chose to do annotation in this way. For example, the three expert reviewers worked without payment, while the two human annotators agreed to work voluntarily during the most inefficient time (trial time which lasted for about two days).

\end{document}